\documentclass[11pt,a4paper]{article}
\usepackage[hyperref]{ranlp2023}
\usepackage{times}
\usepackage{latexsym}

\usepackage{graphicx}
\usepackage{booktabs}
\usepackage{amsmath}
\usepackage{amsfonts}

\usepackage{microtype}

\aclfinalcopy % Uncomment this line for the final submission
\title{A Multi-Strategy Approach for AI-Generated Text Detection\\[0.1cm]}

\author{
Ali Zain \\
{\tt vin.alizain@gmail.com} 
\And
Sareem Farooqui \\
{\tt sareemfarooqui10@gmail.com} 
\AND
Muhammad Rafi \\
{\tt muhammad.rafi@nu.edu.pk} \\ \\
National University of Computer and Emerging Sciences, FAST \\
Karachi, Pakistan
}

\date{}

\begin{document}
\maketitle
\begin{abstract}
This paper presents presents three distinct systems developed for the M-DAIGT shared task on detecting AI generated content in news articles and academic abstracts. The systems includes: (1) A fine-tuned RoBERTa-base classifier, (2) A classical TF-IDF + Support Vector Machine (SVM) classifier , and (3) An Innovative ensemble model named Candace, leveraging probabilistic features extracted from multiple Llama-3.2 models processed by a custom Transformer encoder.

The RoBERTa-based system emerged as the most performant, achieving near-perfect results on both development and test sets.
\end{abstract}

\section{Introduction}
The proliferation of sophisticated large language models (LLMs) has led to a surge in AI-generated text, making its detection a critical area of research \cite{jawahar-etal-2020-automatic}. Identifying machine-generated content is crucial for maintaining information integrity, combating misinformation \cite{pan-etal-2023-risk}, and ensuring academic honesty. The M-DAIGT (Multi-domain DAIGT) shared task \cite{lamsiyah2025mdaigt} aims to foster research in this domain by providing datasets for two distinct scenarios: news articles (Subtask 1) and academic abstracts (Subtask 2). Participants are tasked with building systems to classify given texts as either human-written or machine-generated.

In response to this challenge, our team developed and evaluated three different systems:
\begin{enumerate}
    \item \textbf{RoBERTa-based Classifier:} A fine-tuned RoBERTa-base model, a widely successful approach for text classification tasks.
    \item \textbf{TF-IDF + SVM Classifier:} A traditional machine learning pipeline combining Term Frequency-Inverse Document Frequency (TF-IDF) features with a Linear Support Vector Machine (SVM) \cite{joachims1998text}. This served as a strong baseline, particularly for Subtask 1.
    \item \textbf{Llama-Feature Ensemble with Transformer Classifier (Candace):} An experimental system designed to capture nuanced signals from multiple LLMs. It extracts probabilistic features \cite{sarvazyan-etal-2024-genaios} (e.g., token log-probabilities, entropy) from a suite of Llama-3.2 models \cite{meta-llama3-2} and uses a custom Transformer Encoder-based model for final classification.
\end{enumerate}
This paper details the architecture, data handling, implementation, and experimental results of these systems on the provided test datasets. Our RoBERTa-based approach yielded the most consistent and high-performing results on the development and test sets and was selected for our final submissions for both subtasks.

\section{System Architectures}
We developed three distinct systems, each employing a different strategy for AI-generated text detection.

\subsection{System 1: RoBERTa-based Classifier}
\begin{figure}[ht]
    \centering
    \includegraphics[width=0.9\columnwidth]{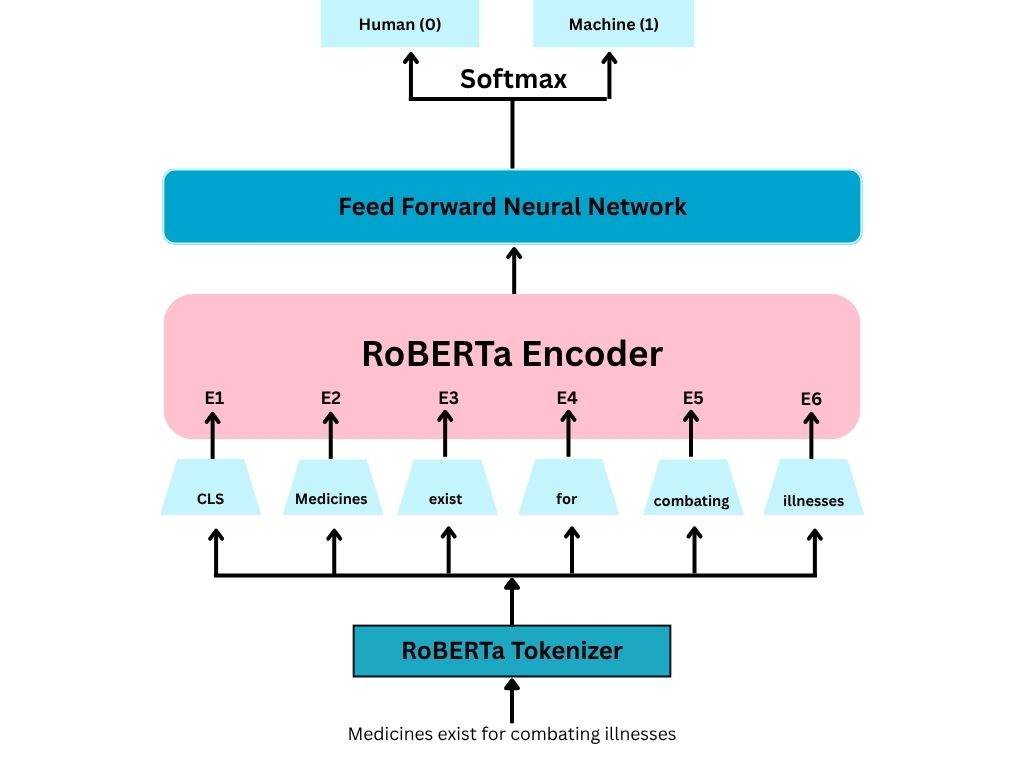} % Placeholder
    \caption{Architecture of System 1: RoBERTa-based Classifier.}
    \label{fig:roberta_arch}
\end{figure}

This system (Figure \ref{fig:roberta_arch}) fine-tunes a pre-trained RoBERTa-base model \cite{liu-etal-2019-roberta}. The input text is tokenized, and the RoBERTa model processes these tokens. The final hidden state corresponding to the special `[CLS]` token is then passed through a linear classification layer to produce a binary prediction (human or machine).

\subsection{System 2: TF-IDF + SVM Classifier}
\begin{figure}[ht]
    \centering
    \includegraphics[width=0.9\columnwidth]{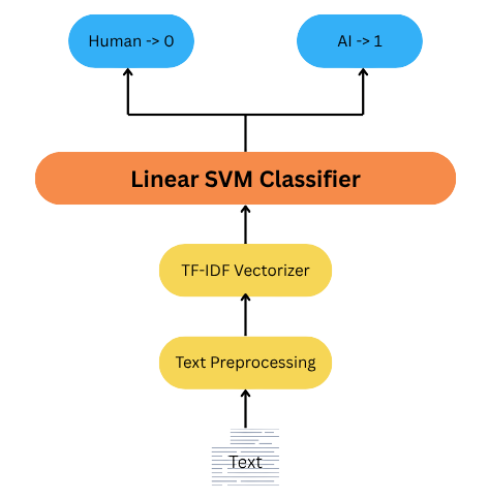} % Placeholder
    \caption{Architecture of System 2: TF-IDF + SVM Classifier.}
    \label{fig:tfidf_svm_arch}
\end{figure}

Our second system (Figure \ref{fig:tfidf_svm_arch}) follows a traditional machine learning pipeline. Textual input is first converted into a numerical representation using TF-IDF vectorization, capturing n-grams. These TF-IDF features are then fed into a Linear Support Vector Machine (SVM) for classification.

\subsection{System 3: Llama-Feature Ensemble with Transformer Classifier (Candace)}
\begin{figure}[ht]
    \centering
    \includegraphics[width=0.9\columnwidth]{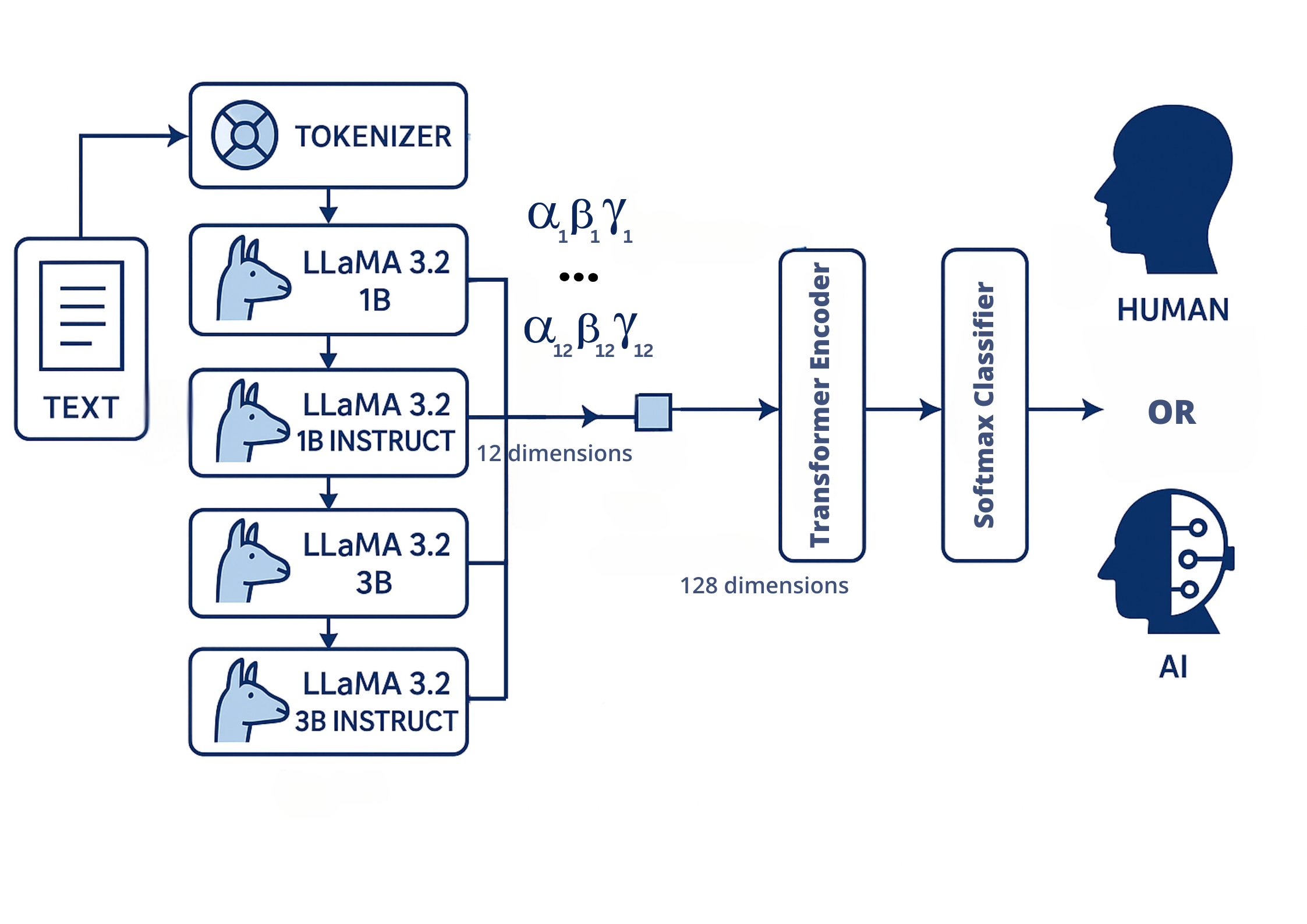} % Placeholder
    \caption{Architecture of System 3: Candace - Llama-Feature Ensemble with Transformer Classifier.}
    \label{fig:candace_arch}
\end{figure}

The third system (Figure \ref{fig:candace_arch}), named Candace, is more experimental. It involves a two-stage process. First, probabilistic features (alpha, beta, gamma, as described in Section \ref{sec:method_candace}) are extracted from each token of the input text using multiple Llama-3.2 models. These feature vectors are concatenated. Second, this sequence of combined Llama-derived features is processed by a custom Transformer Encoder-based classification head, which then makes the final human/machine prediction.

\section{Data and Resources}
The M-DAIGT shared task provided datasets for two subtasks:
\begin{itemize}
    \item \textbf{Subtask 1 (News):} Comprised of `T1\_train.csv` (10,000 samples), `T1\_dev.csv` (2,000 samples), and `T1\_test\_unlabeled.csv` (2,000 samples).
    \item \textbf{Subtask 2 (Academic Abstracts):} Comprised of `T2\_train.csv` (10,000 samples), `T2\_dev.csv` (2,000 samples), and `T2\_test\_unlabeled.csv` (2,000 samples).
\end{itemize}
Each labeled dataset contained an `id`, `text`, and a `label` column, where labels were either `human` or `machine`. For training, labels were mapped to 0 (human) and 1 (machine).
Minimal preprocessing was applied for the RoBERTa and TF-IDF systems, primarily consisting of standard tokenization handled by the respective libraries. The Candace system's feature extraction used raw text.
External resources included:
\begin{itemize}
    \item Pre-trained `roberta-base` model and tokenizer from Hugging Face Transformers \cite{wolf-etal-2020-transformers}.
    \item Pre-trained Llama-3.2 models \cite{meta-llama3-2} (`meta-llama/Llama-3.2-1B`, `meta-llama/Llama-3.2-1B-Instruct`, `meta-llama/Llama-3.2-3B`, `meta-llama/Llama-3.2-3B-Instruct`) and the `meta-llama/Llama-3.2-1B` tokenizer.
\end{itemize}

\section{Methodology}

\subsection{System 1: RoBERTa-based Classifier}
\paragraph{Model Architecture:} We used the `RobertaModel` from Hugging Face Transformers, pre-trained on `roberta-base`. A linear classification layer was added on top of the pooled output (representation of the `[CLS]` token) from the RoBERTa model. The output layer predicts a score for the two classes (human vs. machine).
\paragraph{Input Representation:} Texts were tokenized using `RobertaTokenizerFast` with a maximum sequence length of 512 tokens. Padding was applied to shorter sequences, and longer sequences were truncated.
\paragraph{Training:} The model was fine-tuned for 4 epochs using the Adam optimizer with a learning rate of $1 \times 10^{-5}$. We used `CrossEntropyLoss` as the loss function. The batch size was set to 16. This setup was applied independently for both Subtask 1 and Subtask 2, using their respective training and development datasets.

\subsection{System 2: TF-IDF + SVM Classifier}
This system was developed primarily as a baseline for Subtask 1 (News).
\paragraph{Feature Extraction:} We used \href{https://scikit-learn.org/stable/modules/generated/sklearn.feature_extraction.text.TfidfVectorizer.html}{\texttt{TfidfVectorizer}} from scikit-learn to convert text into numerical features. We configured it to use n-grams of range (2, 3) and limited the maximum number of features to 5,000.

\paragraph{Classifier:} A Linear Support Vector Machine (LinearSVC) was employed for classification. The hyperparameters were set as follows: C (regularization parameter) = 0.5, `class\_weight='balanced'` to handle potential class imbalance, `dual=False` (as n\_samples were greater than n\_features), and `max\_iter=5000` to ensure convergence.

\subsection{System 3: Llama-Feature Ensemble with Transformer Classifier (Candace)}
\label{sec:method_candace}
This experimental system explores the utility of probabilistic features derived from multiple instruction-tuned and base Llama-3.2 models.
\paragraph{Feature Extraction:} For each input text and for each of the four Llama models (`meta-llama/Llama-3.2-1B`, `meta-llama/Llama-3.2-1B-Instruct`, `meta-llama/Llama-3.2-3B`, `meta-llama/Llama-3.2-3B-Instruct`), we extracted three features per token up to a maximum sequence length of 256:
\begin{itemize}
    \item \textbf{Alpha ($\alpha$):} The maximum log-probability assigned by the Llama model to any token at that position, given the preceding tokens.
    \item \textbf{Beta ($\beta$):} The entropy of the Llama model's predicted probability distribution over the vocabulary at that position.
    \item \textbf{Gamma ($\gamma$):} The log-probability assigned by the Llama model to the actual observed token at that position.
\end{itemize}
The Llama models were loaded with 8-bit quantization to manage memory. Features from all four Llama models were concatenated token-wise, resulting in $4 \text{ models} \times 3 \text{ features from each model} = 12$ features per token.
\paragraph{Classifier Architecture (CandaceClassifier):} The sequence of aggregated indicators was then processed by a custom classification architecture. This architecture begins with a projection of the indicator sequence into a higher-dimensional space. This transformed sequence is then passed through a Transformer Encoder block, designed to capture contextual relationships between the token-level indicators. The output of the Transformer Encoder is subsequently pooled across the sequence dimension, and a final linear layer produces the binary classification.
\paragraph{Training:} The CandaceClassifier was trained for 10 epochs using AdamW optimizer \cite{loshchilov-hutter-2019-adamw} with a learning rate of $1 \times 10^{-4}$ and `CrossEntropyLoss`. The batch size was 8. This architecture was trained separately for Subtask 1 and Subtask 2.

\section{Experiments and Results}
All systems were trained and evaluated on the M-DAIGT development sets for their respective subtasks. The primary evaluation metrics were Accuracy and F1-score.

\begin{table*}[ht]
\centering
\small
\begin{tabular}{@{}lcccc|cccc@{}}
\toprule
 & \multicolumn{4}{c}{\textbf{Subtask 1 (News) - Test Set}} & \multicolumn{4}{c}{\textbf{Subtask 2 (Academic) - Test Set}} \\
\textbf{System} & Acc. & F1 & Prec. & Rec. & Acc. & F1 & Prec. & Rec. \\
\midrule
RoBERTa-base (System 1) & \textbf{99.99\%} & \textbf{99.99\%} & \textbf{99.80\%} & \textbf{100.0\%} & \textbf{100.00\%} & \textbf{100.00\%} & \textbf{100.00\%} & \textbf{100.00\%} \\
TF-IDF + SVM (System 2) & 97.90\% & 97.91\% & 97.52\% & 98.30\% & 99.85\% & 99.85\% & 100.0\%& 99.70\% \\
Candace (System 3) & 99.75\% & 99.75\% & 99.60\% & 99.90\% & 99.95\%$^\dagger$ & 99.95\% & 100.00\% & 99.90\% \\
\bottomrule
\end{tabular}
\caption{Test set performance, RoBERTa base model with Fast tokenizer outperforming all models}
\label{tab:results}
\end{table*}

\paragraph{RoBERTa-based System (System 1):}
For Subtask 1 (News), our fine-tuned RoBERTa model achieved an accuracy of 99.95\% and an F1-score of 99.95\% on the development set (best at epoch 4).
For Subtask 2 (Academic Abstracts), the RoBERTa model achieved 100.00\% accuracy and 100.00\% F1-score on the development set (stable from epoch 1 onwards).
Given its strong and consistent performance, this system was chosen for our official submissions for both subtasks.

\paragraph{TF-IDF + SVM System (System 2):}
This system was evaluated on Subtask 1 and Subtask 2. On its internal training data (as dev metrics were not explicitly separated in its notebook), it achieved an accuracy of 99.81\% and an F1-score of 0.9981. While competitive, it was slightly outperformed by the RoBERTa model on the development set.

\paragraph{Candace System (System 3):}
For Subtask 1 (News), the Candace system achieved a development accuracy of 99.80\% (best at epoch 6). The same architectural design and training procedure were applied to Subtask 2, and similar development accuracy (99.80\%) was observed during its separate training run. While promising, this system is more computationally intensive due to the multi-LLM feature extraction step. The RoBERTa system offered slightly better or comparable performance with significantly less overhead for these specific datasets.

\section{Discussion}
Our experiments highlight the continued effectiveness of fine-tuned transformer models like RoBERTa for text classification tasks, achieving near-perfect scores on the development sets for both news and academic abstract domains. The RoBERTa model's ability to capture subtle linguistic cues makes it highly suitable for distinguishing between human and AI-generated text.

The TF-IDF + SVM approach, while simpler, provided a very strong baseline for Subtask 1, underscoring the utility of traditional methods, especially when coupled with robust feature engineering like n-grams.

The Candace system, which extracts features from multiple Llama-3.2 models, also showed excellent performance. This approach is interesting as it attempts to distill knowledge from several powerful LLMs into a smaller, specialized classifier. However, the feature extraction process is computationally expensive. For the M-DAIGT datasets, the gains over a well-tuned RoBERTa model were not substantial enough to justify the additional complexity and computational cost as the primary submission.

Runtime for RoBERTa inference is efficient, while Candace inference is slower due to the initial pass through multiple Llama models.

\section{Conclusion}
We presented three distinct systems for detecting AI-generated text in news articles and academic abstracts. Our fine-tuned RoBERTa-base model demonstrated exceptional performance on the development and test sets for both subtasks, achieving near-perfect accuracy and F1-scores, and was selected as our primary submission. The TF-IDF+SVM system served as a strong baseline, and the experimental Candace system, leveraging features from multiple Llama models, also showed high efficacy.
Future work could involve ensembling these diverse models, exploring more sophisticated feature fusion techniques for the Candace system, and investigating the robustness of these models against adversarial attacks or text generated by newer, more advanced language models.

\section*{Acknowledgments}
We thank the organizers of the M-DAIGT shared task for providing the dataset and the evaluation platform. BEsides that, the research is also supported by the provisional award under the National Research Program for Universities (NRPU), Higher Education Commission (HEC) Pakistan, with the title “NRPU: Automatic Multi-Model Classification of Religious Hate Content from Social Media” (Reference Research Project No. 16153).

% Entries for the entire Anthology, followed by custom entries
\bibliography{anthology,ranlp2023}
\bibliographystyle{acl_natbib}

\appendix
% Appendix sections are not counted towards the page limit.
% \section{Example Appendix}
% \label{sec:appendix}
% This is an appendix.

\end{document}